\documentclass[10pt,twocolumn,letterpaper]{article}

\usepackage{cvpr}              % To produce the CAMERA-READY version
\usepackage{graphicx}
\usepackage{amsmath}
\usepackage{amssymb}
\usepackage{booktabs}
\usepackage{multirow}
\usepackage{pifont}
\usepackage{accsupp}
\usepackage[pagebackref,breaklinks,colorlinks]{hyperref}

\usepackage[capitalize]{cleveref}
\crefname{section}{Sec.}{Secs.}
\Crefname{section}{Section}{Sections}
\Crefname{table}{Table}{Tables}
\crefname{table}{Tab.}{Tabs.}

\def\cvprPaperID{5618} % *** Enter the CVPR Paper ID here
\def\confName{CVPR}
\def\confYear{2023}

\begin{document}

\title{A New Comprehensive Benchmark for \\ Semi-supervised Video Anomaly Detection and Anticipation}
\author{Congqi Cao\textsuperscript{\textdagger} \and Yue Lu \and Peng Wang \and Yanning Zhang \\
\and ASGO, School of Computer Science, Northwestern Polytechnical University, China\\
{\tt\small congqi.cao@nwpu.edu.cn zugexiaodui@mail.nwpu.edu.cn \{peng.wang, ynzhang\}@nwpu.edu.cn}
%\and
}
\maketitle
\footnotetext[2]{Corresponding author}

%%%%%%%%% ABSTRACT
\begin{abstract}
Semi-supervised video anomaly detection (VAD) is a critical task in the intelligent surveillance system.
However, an essential type of anomaly in VAD named scene-dependent anomaly has not received the attention of researchers.
Moreover, there is no research investigating anomaly anticipation, a more significant task for preventing the occurrence of anomalous events.
To this end, we propose a new comprehensive dataset, NWPU Campus, containing 43 scenes, 28 classes of abnormal events, and 16 hours of videos.
At present, it is the largest semi-supervised VAD dataset with the largest number of scenes and classes of anomalies, the longest duration, and the only one considering the scene-dependent anomaly.
Meanwhile, it is also the first dataset proposed for video anomaly anticipation.
We further propose a novel model capable of detecting and anticipating anomalous events simultaneously.
Compared with 7 outstanding VAD algorithms in recent years, our method can cope with scene-dependent anomaly detection and anomaly anticipation both well, achieving state-of-the-art performance on ShanghaiTech, CUHK Avenue, IITB Corridor and the newly proposed NWPU Campus datasets consistently.
Our dataset and code is available at: \url{https://campusvad.github.io}.

\end{abstract}

\section{Introduction}
Video anomaly detection (VAD) is widely applied in public safety and intelligent surveillance due to its ability to detect unexpected abnormal events in videos.
Since anomalous events are characterized by unbounded categories and rare occurrence in practice, VAD is commonly set as a semi-supervised task, that is, there are only normal events without specific labels in the training \textcolor{black}{set} \cite{AnomalyDetection2009chandola, SurveySingleScene2022ramachandra}.
The model trained only on the normal events needs to distinguish anomalous events from normal events in the testing phase.

Semi-supervised VAD has been studied for years.
Especially in recent years, reconstruction-based \textcolor{black}{and prediction-based} methods \cite{FutureFrame2018liu, MemorizingNormality2019gong, AnomalyDetection2019nguyen, LearningMemoryGuided2020park, LearningNormal2020song, NormalityLearning2020zhang, BMANBidirectional2020lee, HybridVideo2021liu, LearningNormal2021lv, AttentionDrivenLoss2020zhou, ClozeTest2020yu, AbnormalEvent2021yu, FutureFrame2021luo, NMGANNoisemodulated2021chen, RobustUnsupervised2021wang, MultiEncoderEffective2021fang, AnomalyDetection2022fang, InfluenceawareAttention2022zhang, VariationalAbnormal2022li} have made leaps and bounds in performance on existing datasets.
For example, the frame-level AUCs (area under curve) on UCSD Ped1 and Ped2 datasets \cite{AnomalyDetection2010mahadevan} have reached over 97\% \cite{SurveySingleScene2022ramachandra}.
Despite the emergence of a few challenging datasets, researchers still overlook an important type of anomaly, \ie, the scene-dependent anomaly \cite{SurveySingleScene2022ramachandra}.
Scene dependency refers to that an event is normal in one scene but abnormal in another.
For example, playing football on the playground is a normal behavior, but playing on the road is abnormal.
Note that single-scene datasets cannot contain any scene-dependent anomaly.
Nevertheless, the existing multi-scene datasets (\eg, ShanghaiTech \cite{RevisitSparse2017luo}, UBnormal \cite{UBnormalNew2022acsintoae}) also have not taken this type of anomaly into account.
As a result, there is currently no algorithm for studying scene-dependent anomaly detection, limiting the comprehensive evaluation of VAD algorithms.
In addition to detecting various types of anomalies, we argue that there is another task that also deserves the attention of researchers, which is to anticipate the occurrence of abnormal events in advance.
If we can make an early warning before the anomalous event occurs based on the trend of the event, it is of great significance to prevent dangerous accidents and avoid loss of life and property.
However, according to our investigation, there is no research on video anomaly anticipation, and no dataset or algorithm has been proposed for this field.

\begin{table*}[!t]
	\centering
	\caption{Comparisons of different semi-supervised VAD datasets. There are not any official training and testing splits in UMN. UBnormal has a validation set, which is not shown here. "720p" means that the frame is 720 pixels high and 1280 or 1080 pixels wide. The frame resolutions of NWPU Campus are 1920$\times$1080, 2048$\times$1536, 704$\times$576 and 1280$\times$960 pixels. * represents the animated dataset.}
	\label{tab1}
	\setlength\tabcolsep{5.5pt}
	\begin{tabular}{@{}lcrrrcccc@{}}
		\toprule
		\multicolumn{1}{c}{\multirow{2}{*}{Dataset}} & \multirow{2}{*}{Year} & \multicolumn{3}{c}{\# Frames}    & \multirow{2}{*}{\begin{tabular}[c]{@{}c@{}}\# Abnormal\\ event classes\end{tabular}} & \multirow{2}{*}{Resolution} & \multirow{2}{*}{\#Scenes} & \multirow{2}{*}{\begin{tabular}[c]{@{}c@{}}Scene\\ dependency\end{tabular}} \\ \cmidrule(lr){3-5}
		\multicolumn{1}{c}{}                         &                       & Total     & Training  & Testing &                                                                            &                             &                          &                                                                             \\ \midrule
		Subway Entrance \cite{RobustRealTime2008adam}                         & 2008                  & 86,535    & 18,000    & 68,535  & 5                                                                          & 512$\times$384              & 1                        & \ding{55}                                                                   \\
		Subway Exit \cite{RobustRealTime2008adam}                               & 2008                  & 38,940    & 4,500     & 34,440  & 3                                                                          & 512$\times$384              & 1                        & \ding{55}                                                                   \\
		UMN	\cite{CrowdActivityAllAvi}	                                 & 2009                  & 7,741     & -    	 & -  	   & 3 																			& 320$\times$240      		  & 3    					 & \ding{55}                                                                   \\
		USCD Ped1 \cite{AnomalyDetection2010mahadevan}                              	 & 2010                  & 14,000    & 6,800     & 7,200   & 5                                                                          & 238$\times$158              & 1                        & \ding{55}                                                                   \\
		USCD Ped2 \cite{AnomalyDetection2010mahadevan}                                 & 2010                  & 4,560     & 2,550     & 2,010   & 5                                                                          & 360$\times$240              & 1                        & \ding{55}                                                                   \\
		CUHK Avenue \cite{AbnormalEvent2013lu}                               & 2013                  & 30,652    & 15,328    & 15,324  & 5                                                                          & 640$\times$360              & 1                        & \ding{55}                                                                   \\
		ShanghaiTech \cite{RevisitSparse2017luo}                                & 2017                  & 317,398   & 274,515   & 42,883  & 11                                                                          & 856$\times$480              & 13                       & \ding{55}                                                                   \\
		Street Scene \cite{StreetScene2020ramachandra}                              & 2020                  & 203,257   & 56,847    & 146,410 & 17                                                                         & 1280$\times$720~~           & 1                        & \ding{55}                                                                   \\
		IITB Corridor \cite{MultitimescaleTrajectory2020rodrigues}                               & 2020                  & 483,566   & 301,999   & 181,567 & 10                                                                         & 1920$\times$1080            & 1                        & \ding{55}                                                                   \\
		UBnormal \cite{UBnormalNew2022acsintoae} *                                    & 2022                  & 236,902   & 116,087   & 92,640  & 22                                                                         & 720p                        & 29                       & \ding{55}                                                                   \\
		NWPU Campus                                & (ours)                & \textbf{1,466,073} & \textbf{1,082,014} & \textbf{384,059} & \textbf{28}                                                                         & multiple                    & \textbf{43}                       & \ding{51}                                                                   \\ \bottomrule
	\end{tabular}
\end{table*}

\begin{figure*}[!t]
	\centering
	\includegraphics[]{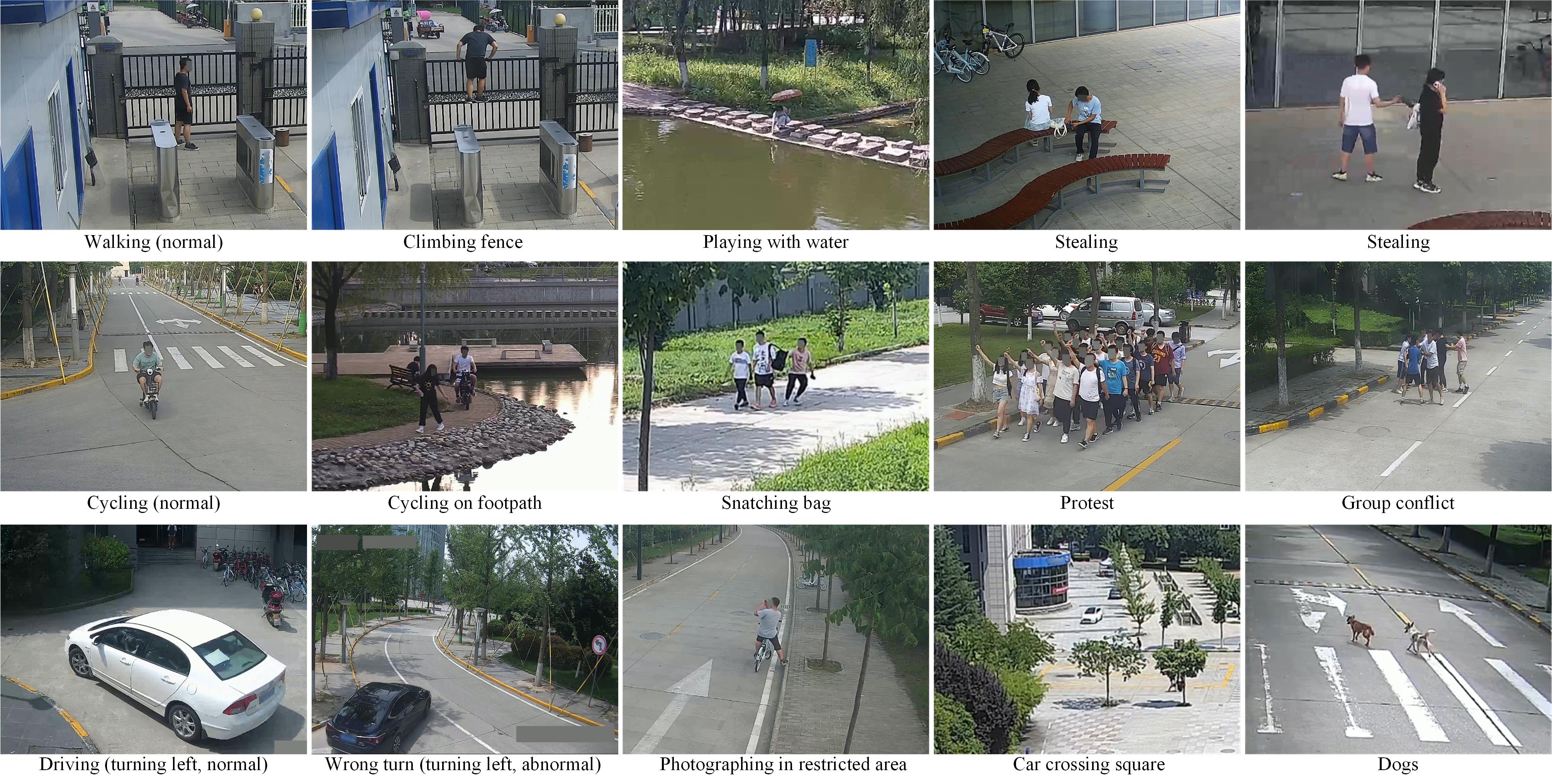}
	\caption{Samples from the proposed NWPU Campus dataset. The samples in the first column are normal events, while the others are different types of anomalous events.}
	\label{fig1}
\end{figure*}

In this paper, we work on semi-supervised video anomaly detection and anticipation.
First and foremost, to address the issue that the VAD datasets lack scene-dependent anomalies and are not suitable for anomaly anticipation, we propose a new large-scale dataset, NWPU Campus.
Compared with existing datasets, our proposed dataset mainly has the following three advantages.
First, to the best of our knowledge, the NWPU Campus is the largest semi-supervised VAD dataset to date.
It contains 43 scenes, whose number is 3 times that of ShanghaiTech, the real recorded dataset with the largest number of scenes among the existing datasets.
The total video duration of the NWPU Campus is 16 hours, which is more than 3 times that of the existing largest semi-supervised VAD dataset IITB Corridor \cite{MultitimescaleTrajectory2020rodrigues}.
The quantitative comparison between the NWPU Campus and other datasets can be seen in \cref{tab1}.
Second, the NWPU Campus has a variety of abnormal and normal events.
In terms of anomalies, it contains 28 classes of anomalous events, which is more than any other dataset.
\cref{fig1} displays some examples from our dataset.
More importantly, the NWPU Campus dataset contains scene-dependent anomalous events, which are missing in other datasets.
As an example, the behavior of a vehicle turning left is anomalous in the scene where left turns are prohibited, while it is normal in other unrestricted scenes.
Along with the diversity of anomalous events, the normal events in our dataset are diverse as well.
Unlike other datasets, we do not only take walking and standing as normal behaviors.
In our dataset, regular walking, cycling, driving and other daily behaviors that obey rules are also considered as normal events.
Third, in addition to being served as a video anomaly detection benchmark, the NWPU Campus is the first dataset proposed for video anomaly anticipation (VAA).
The existing datasets do not deliberately consider the anomalous events applicable to anticipation.
In contrast, we take into account the complete process of the events in the data collection phase so that the occurrence of abnormal events is predictable. 
%As the whole process of the event is captured in our dataset, it is possible for algorithms to anticipate whether an anomalous event will occur or not according to the trend of the event.
For instance, before the vehicle turns left (the scene-dependent anomalous event as mentioned before), the movement trend of it can be observed, and hence the algorithm could make an early warning.
As a comparison, it is considered to be abnormal when a vehicle simply appears in the ShanghaiTech dataset, which is unpredictable and therefore not suitable for anomaly anticipation.

Besides comprehensive benchmarks, there is currently a lack of algorithms for scene-dependent anomaly detection and video anomaly anticipation.
Therefore, in this work, we further propose a novel forward-backward frame prediction model that can detect anomalies and simultaneously anticipate whether an anomalous event is likely to occur in the future.
Moreover, it has the ability to handle scene-dependent anomalies through the proposed scene-conditioned auto-encoder.
As a result, our method achieves state-of-the-art performance on ShanghaiTech \cite{RevisitSparse2017luo}, CUHK Avenue \cite{AbnormalEvent2013lu}, IITB Corridor \cite{MultitimescaleTrajectory2020rodrigues}, and our NWPU Campus datasets.

In summary, our contribution is threefold:

\begin{itemize}
	\item We propose a new dataset NWPU Campus, which is the largest and most complex semi-supervised video anomaly detection benchmark to date. It makes up for the lack of scene-dependent anomalies in the current research field.
	\item We propose a new video anomaly anticipation task to anticipate the occurrence of anomalous events in advance, and the NWPU Campus is also the first dataset proposed for anomaly anticipation, filling the research gap in this area.
	\item We propose a novel method to detect and anticipate anomalous events simultaneously, and it can cope with scene-dependent anomalies. Comparisons with 7 state-of-the-art VAD methods on the NWPU Campus, ShanghaiTech, CUHK Avenue and IITB Corridor datasets demonstrate the superiority of our method.
	
\end{itemize}

\section{Related Work}
\subsection{Video Anomaly Detection Datasets}
We focus on semi-supervised video anomaly detection in this paper, so the weakly-supervised video anomaly detection datasets such as UCF-Crime \cite{RealWorldAnomaly2018sultani} and XD-Violence \cite{NotOnly2020wu} will not be discussed.
The commonly used semi-supervised VAD datasets include USCD Ped1 \& Ped2 \cite{AnomalyDetection2010mahadevan}, Subway Entrance \& Exit \cite{RobustRealTime2008adam}, UMN \cite{CrowdActivityAllAvi}, CUHK Avenue \cite{AbnormalEvent2013lu}, ShanghaiTech \cite{RevisitSparse2017luo}, Street Scene \cite{StreetScene2020ramachandra}, IITB Corridor \cite{MultitimescaleTrajectory2020rodrigues} and UBnormal \cite{UBnormalNew2022acsintoae}.

The UCSD Ped1 \& Ped2 \cite{AnomalyDetection2010mahadevan} datasets each contain a camera overlooking a pedestrian walkway, in which most of the anomalies are intrusions of other objects, such as bicycles, cars and skateboards.
Therefore, the anomalies can be readily detected through static images, resulting in the saturation of performance (97.4\% \cite{AbnormalEvent2017ravanbakhsh} on Ped1 and 99.2\% \cite{RobustAnomaly2019vu} on Ped2 in frame-level AUC).
The Subway Entrance \& Exit \cite{RobustRealTime2008adam} datasets include two indoor scenes of the subway entrance and exit. The abnormal events are only related to people, including jumping through turnstiles, wrong direction, \etc.
The UMN \cite{CrowdActivityAllAvi} contains three outdoor scenes, and the only type of anomalous event is the crowd dispersing suddenly.
There are not any official training and testing splits in it.
The CUHK Avenue \cite{AbnormalEvent2013lu} contains a camera looking at the side of a building with pedestrian walkways by it, and the abnormal behaviors include running, throwing bags, child skipping, \etc.
The ShanghaiTech \cite{RevisitSparse2017luo} includes a total of 13 outdoor scenes on the campus, and quite a few of the anomalous events are related to objects, such as bicycles, cars, skateboards and strollers, even though it seems normal for these objects to appear in real life.
The anomalous events in ShanghaiTech are generic across scenes and this dataset does not contain scene-dependent anomalies.
The Street Scene \cite{StreetScene2020ramachandra} contains a camera looking down on a scene of a two-lane street with bike lanes and pedestrian sidewalks.
Compared with previous datasets, it includes location anomalies, such as cars parked illegally and cars outside a car lane.
The IITB Corridor \cite{MultitimescaleTrajectory2020rodrigues} is the largest single-scene semi-supervised VAD dataset as far as we know.
The scene consists of a corridor where the normal activities are walking and standing, and the anomalous behaviors are performed by volunteers, including chasing, fighting, playing with ball, \etc.
The UBnormal \cite{UBnormalNew2022acsintoae} is generated by animations, containing a total of 22 types of abnormal events in 29 virtual scenarios.
However, there is a distribution gap between \textcolor{black}{animated} videos and real recorded videos.
Acsintoae \etal \cite{UBnormalNew2022acsintoae} have to use an additional model (CycleGAN \cite{UnpairedImagetoImage2017zhu}) to reduce the distribution gap.

It should be noted that all of the above datasets do not take scene-dependent anomalies and anomaly anticipation into account.
Therefore, a benchmark for the comprehensive evaluation of anomaly detection and anticipation is pressingly needed in the current research stage.
The proposed NWPU Campus dataset has the features of large scale, multiple scenarios, and diverse as well as extensive events.
It is committed to meeting the requirement for a new comprehensive benchmark.

\subsection{Video Anomaly Detection Methods}

Prevalent semi-supervised VAD methods mainly contain distance-based \cite{LearnableLocalitySensitive2022lu, ClusteringDriven2020chang, DeepOneClass2020wu}, reconstruction-based \cite{LearningTemporal2016hasan, AnomalyDetection2019nguyen, MemorizingNormality2019gong, OldGold2020zaheer} and \textcolor{black}{prediction-based} \cite{FutureFrame2018liu, LearningMemoryGuided2020park, LearningNormal2020song, NormalityLearning2020zhang, BMANBidirectional2020lee, HybridVideo2021liu, LearningNormal2021lv, AttentionDrivenLoss2020zhou, ClozeTest2020yu, AbnormalEvent2021yu, FutureFrame2021luo, NMGANNoisemodulated2021chen, RobustUnsupervised2021wang, MultiEncoderEffective2021fang, AnomalyDetection2022fang, InfluenceawareAttention2022zhang, VariationalAbnormal2022li} methods.
Especially, the prediction-based methods have attracted \textcolor{black}{wide} attention in recent years.
They usually predict the current (\ie, the last observed) frame \textcolor{black}{via} previous frames \cite{FutureFrame2018liu, LearningMemoryGuided2020park, LearningNormal2020song, NormalityLearning2020zhang, HybridVideo2021liu, LearningNormal2021lv, AttentionDrivenLoss2020zhou, ClozeTest2020yu, AbnormalEvent2021yu, FutureFrame2021luo, NMGANNoisemodulated2021chen, RobustUnsupervised2021wang, MultiEncoderEffective2021fang, InfluenceawareAttention2022zhang, VariationalAbnormal2022li}, and compute anomaly score based on the error between the predicted frame and the observable groundtruth frame.
To discriminate between abnormal and normal motion, some methods \cite{VariationalAbnormal2022li, HybridVideo2021liu} use optical flow as the condition of conditional VAE to enhance frame prediction.
Combining with memory modules \cite{MemorizingNormality2019gong, LearningMemoryGuided2020park, LearningNormal2021lv, AppearanceMotionMemory2021caia, HybridVideo2021liu} that can explicitly utilize normal patterns is also an improvement trend of this kind of methods.
Besides predicting \textcolor{black}{the current frame}, there are a few \textcolor{black}{prediction-based} methods completing the middle frame with bidirectional frame prediction \cite{BMANBidirectional2020lee, AnomalyDetection2022fang}, which requires the observation of groundtruth frames in both directions.

Different from those prediction-based models, our forward-backward prediction model does not need to observe future frames during inference.
It can estimate the prediction error of future frames whose groundtruth frames are unavailable, \textcolor{black}{making it able to anticipate anomalies}.

\section{Proposed Dataset}
\subsection{Dataset Collection}
\begin{table}[!t]
	\centering
	\caption{Frame count and duration of the NWPU Campus dataset.}
	\label{tab2}
	\begin{tabular}{|ccc|}
		\hline
		\multicolumn{3}{|c|}{NWPU Campus (25 FPS)}                                                     \\ \hline
		\multicolumn{3}{|c|}{1,466,073 (16.29h)}                                                       \\ \hline
		\multicolumn{1}{|c|}{Training frames}    & \multicolumn{2}{c|}{Testing frames}                 \\ \hline
		\multicolumn{1}{|c|}{1,082,014 (12.02h)} & \multicolumn{2}{c|}{384,059 (4.27h)}                \\ \hline
		\multicolumn{1}{|c|}{Normal}             & \multicolumn{1}{c|}{Normal}         & Abnormal      \\ \hline
		\multicolumn{1}{|c|}{1,082,014 (12.02h)} & \multicolumn{1}{c|}{318,793(3.54h)} & 65,266(0.73h) \\ \hline
	\end{tabular}
\end{table}

We set up cameras at 43 outdoor locations on the campus to record the activities of pedestrians and vehicles.
As anomalous events rarely occur in real life, there are a total of more than 30 volunteers performing a part of normal and abnormal events.
In our dataset, the classes of normal events include regular walking, cycling, driving and other daily behaviors that obey rules.
The types of anomalies consist of single-person anomalies (\eg, climbing fence, playing with water), interaction anomalies (\eg, stealing, snatching bag), group anomalies (\eg, protest, group conflict), scene-dependent anomalies (\eg, cycling on footpath, wrong turn, photographing in restricted area), location anomalies (\eg, car crossing square, crossing lawn), appearance anomalies (\eg, dogs, trucks) and trajectory anomalies (\eg, jaywalking, u-turn).
Some normal and abnormal samples are shown in \cref{fig1}.
There are different manifestations for each kind of anomalous event in our dataset.
For instance, stealing may occur when two people are sitting next to each other or when one person is following another.
Additionally, to avoid algorithms detecting anomalies according to specific performers, the volunteers also perform normal behaviors that are similar to the anomalous behavior if possible.
For example, the normal behavior served as a contrast to climbing fence is merely walking up to the fence and then leaving.

Finally, we collect 16 hours of videos from these 43 scenes, including 305 training videos and 242 testing videos.
In the training data, there are only normal events that come from real events (without volunteers) and performed events (with volunteers), while the testing data contains both normal events and anomalous events.
In the testing set, there are a total of 28 classes of abnormal events, most of which are performed by volunteers and some actually occur.
All the anomaly classes and the anomaly classes in each scene are provided in the supplementary material.
We annotate frame-level labels for the testing videos to indicate the presence or absence of anomalous events in each frame.
According to the setting of semi-supervised VAD, algorithms only need to distinguish abnormality from normality.
Thus, the specific classes of the abnormal events are not annotated.
It should be noted that not all the testing videos contain anomalies, since there is no guarantee that an anomalous event will certainly happen in a video in practical applications.
To protect the privacy of volunteers and pedestrians, all the faces in our dataset are blurred.

\subsection{Dataset Statistics}

%\begin{figure}[!t]
%	\centering
%	\includegraphics[]{fig2.pdf}
%	\caption{Duration distributions of training and testing videos in the NWPU Campus dataset.}
%	\label{fig2}
%\end{figure}
%
%\begin{figure}[!t]
%	\centering
%	\includegraphics[]{fig3.pdf}
%	\caption{Distribution of abnormal testing videos according to the percentage of abnormal frames in each video.}
%	\label{fig3}
%\end{figure}

\begin{figure}[!t]
	\centering
	\includegraphics[]{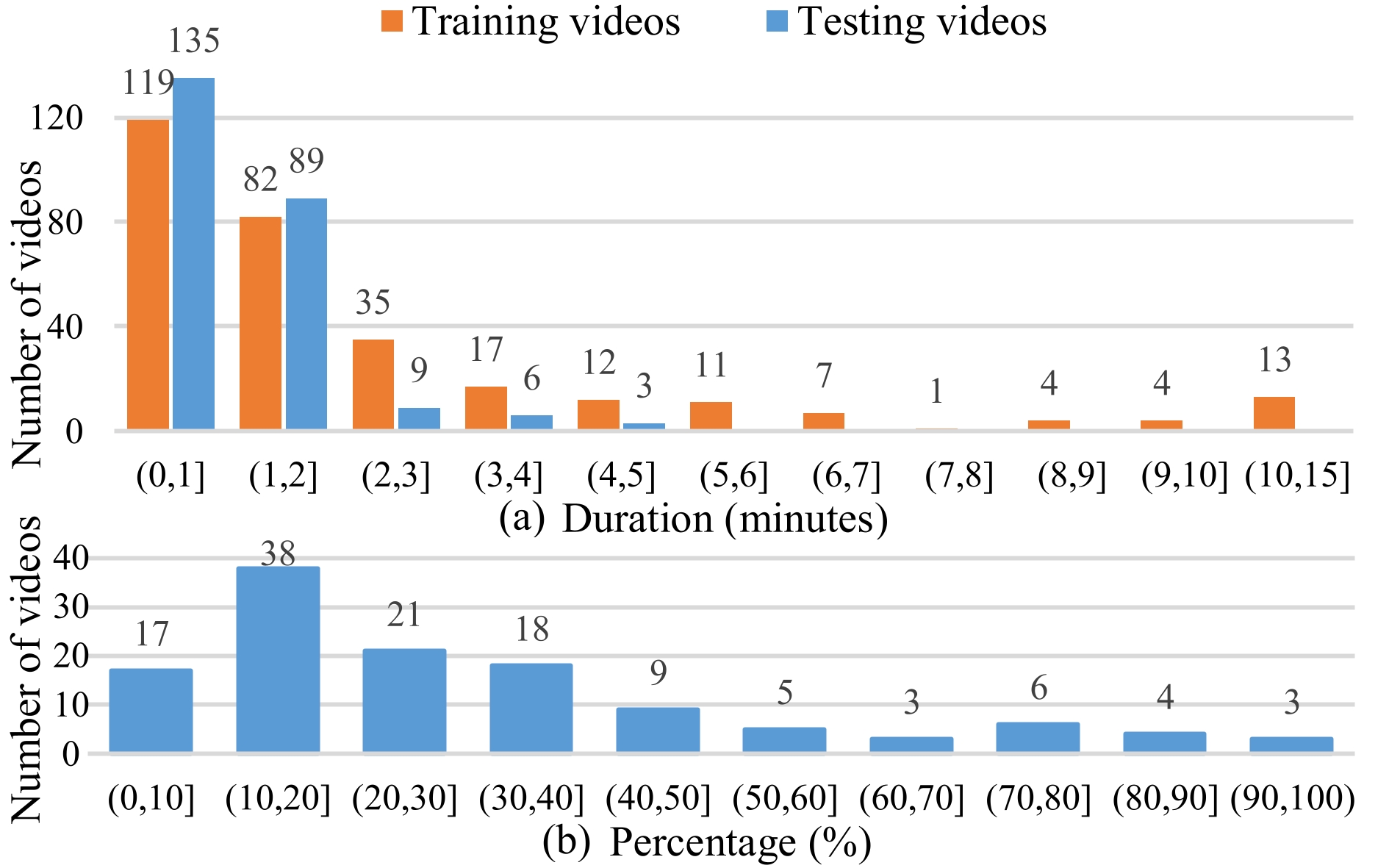}
	\caption{\textcolor{black}{The distributions of training and testing videos according to duration (a), and abnormal testing videos according to the percentage of abnormal frames in each video (b).}}
	\label{fig2}
\end{figure}

The statistics of frame count and duration of our NWPU Campus dataset are shown in \cref{tab2}.
The entire dataset lasts 16.29 hours, involving 4.27 hours of the testing set.
%The frame rate is 25 FPS for all videos, and the entire dataset lasts 16.29 hours, involving 4.27 hours of the testing set.
%The testing videos consist of normal frames and abnormal frames, accounting for 3.54 hours and 0.73 hours, respectively.
\cref{fig2}(a) shows the duration distribution of the 305 training videos and 242 testing videos.
The average duration of the training videos is 2.37 minutes, and that of the testing videos is 1.05 minutes.
There are 124 videos in the testing set that contain anomalous events, and \cref{fig2}(b) presents the percentage of abnormal frames in the abnormal videos.

In order to highlight the traits of our dataset, we comprehensively compare NWPU Campus with other widely-used datasets for semi-supervised video anomaly detection, as shown in \cref{tab1}.
It can be concluded that the proposed NWPU Campus dataset has three outstanding traits.
First, it is the largest semi-supervised video anomaly detection dataset, which is over three times larger than the existing largest dataset (\ie, IITB Corridor).
Second, the scenes and anomaly classes of our dataset are diverse and complex.
It is a real recorded dataset with the largest number of abnormal event classes and scenes by far.
Although the UBnormal dataset also has multiple scenarios, it is a virtual dataset generated by animation rather than real recordings.
Third, our dataset takes into account the scene-dependent anomalous events, which is an important type of anomaly not included in other multi-scene datasets.
Besides the above three advantages, the NWPU Campus is also the first dataset proposed for video anomaly anticipation, which will be introduced in detail in the next section.

\section{Proposed Method}
\subsection{Problem Formulation}

\begin{figure}[!t]
	\centering
	\includegraphics[]{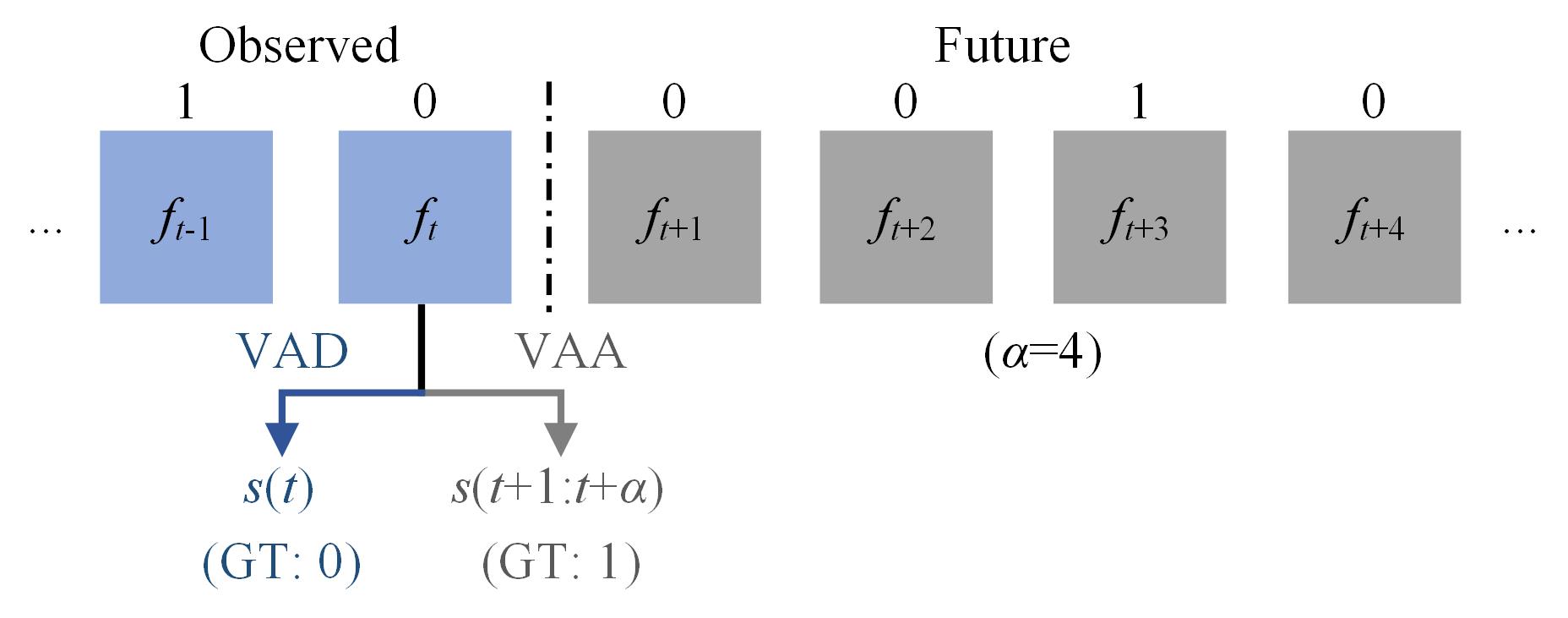}
	\caption{Illustration of video anomaly detection (VAD) and anticipation (VAA). $f_t$ is the frame at time $t$. "0" represents normality and "1" represents abnormality. $s()$ denotes the anomaly score. $\alpha$ is the anticipation time. "GT" stands for groundtruth.}
	\label{fig4}
\end{figure}

Video anomaly detection (VAD) aims to detect whether an anomaly is occurring at the current moment.
As to anomaly anticipation, considering that it is difficult \textcolor{black}{and inessential} to anticipate the exact time of the occurrence of an abnormal event, we define video anomaly anticipation (VAA) to anticipate whether an anomaly will occur in a future period of time, which is meaningful and useful for early warnings of anomalous events.
We illustrate the VAD and VAA tasks in \cref{fig4}.

Suppose the current time step is $t$.
For VAD, an algorithm can compute an anomaly score $s(t)$ for the current frame $f_t$ based on the observed frames $f_{t-n}, \cdots, f_t$, where $n$ represents the observed duration. In \cref{fig4}, $f_t$ is a normal frame, and therefore the anomaly score $s(t)$ is expected to be as low as possible.
For VAA, at the current moment $t$, we anticipate whether an anomaly will occur at any future frame in the period of $[t+1, t+\alpha]$ that has not been observed, where $\alpha \geq 1$ is the anticipation time.
\textcolor{black}{We use the score $s(t+1:t+\alpha)$} to represent the anticipated probability of an anomaly occurring during \textcolor{black}{$t+1$ to $t+\alpha$ frames.}
%The $t+1$ to $t+\alpha$ frames share the same anomaly score to represent the anticipated probability of an anomaly occurring during this period, which is denoted as $s(t+1:t+\alpha)$.
%We denote the anticipated anomaly score as $s(t+1:t+\alpha)$.
In \cref{fig4} where $\alpha=4$ is taken as an example, since $f_{t+3}$ is abnormal, the groundtruth of $s(t+1:t+\alpha)$ is 1, denoting there will be an anomaly in frames $f_{t+1}, \cdots, f_{t+4}$.
We expect that the anomaly score $s(t+1:t+\alpha)$ to be as high as possible, which is contrary to $s(t)$.

As can be seen, the groundtruth is different for VAD and VAA.
For VAD, we denote the frame-level labels of a video as $G_0=\{g_t\}_{t=1}^{T}$, where $g_t \in \{0, 1\}$ indicates the frame $f_t$ is normal (0) or abnormal (1), and $T$ is the length of the video.
Based on $G_0$, the frame-level labels for VAA where the anticipation time is $\alpha$ can be calculated by:
\begin{equation}
	G_{\alpha}=\{\max(\{g_{t+i}\}_{i=1}^{\alpha})\}_{t=1}^{T-\alpha},
\end{equation}
where $\max()$ denotes the maximum value in a set.

Note that the action anticipation models \textcolor{black}{(\eg \cite{AnticipativeVideo2021girdhar, FutureTransformer2022gong, HybridEgocentric2022liu})} are not applicable to semi-supervised VAA, since there are no anomaly data and labels to train them in a supervised manner.
Therefore, we propose a novel method for semi-supervised VAD and VAA in the next section.

\subsection{Forward-backward Scene-conditioned Auto-encoder}
\begin{figure*}[!t]
	\centering
	\includegraphics[]{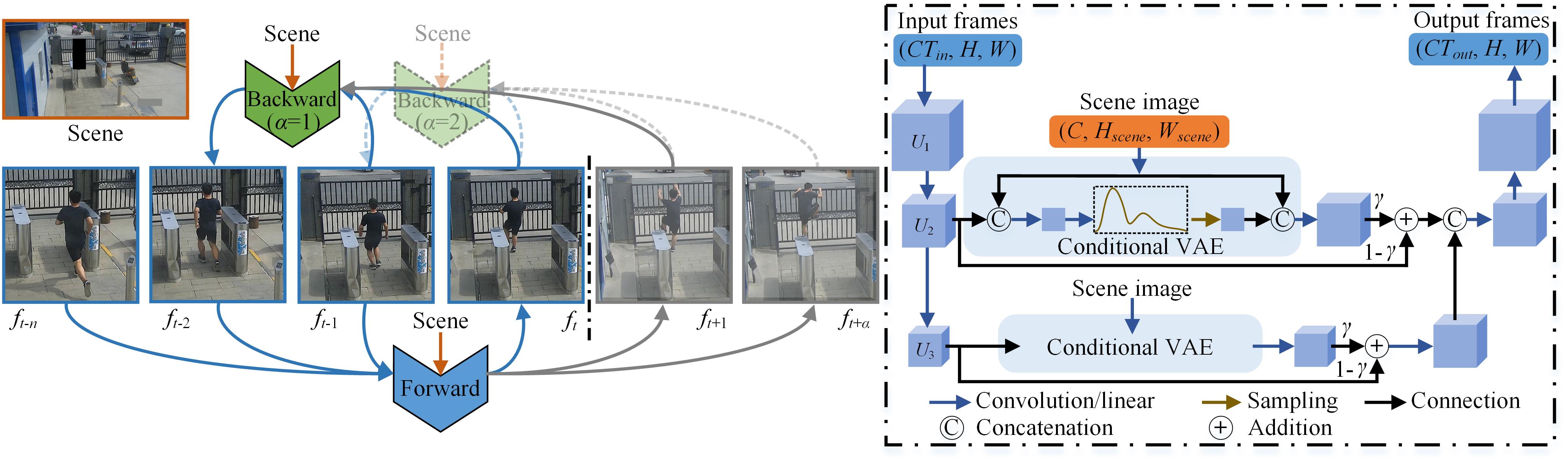}
	\caption{The proposed forward-backward scene-conditioned auto-encoder. It consists of a forward and a backward frame prediction networks. Each network has the same U-Net architecture with conditional VAEs that take the scene image as the input. $t$, $n$ and $\alpha$ respectively represent the current time, the observation time and the anticipation time. $C$, $T$, $H$ and $W$ respectively represent the channel, temporal length, height and width of the input frames. $U_i$ denotes the $i$-th level of the U-Net. $\gamma$ is a weight in scalar.
		Best viewed in color.}
	\label{fig5}
\end{figure*}

%Our model is based on the prevalent frame prediction model, which predicts the current frame based on previous frames.
%The prediction error between the predicted frame and the groundtruth frame reveals the probability of anomaly.
Our model is based on the prevalent frame prediction model.
However, future groundtruth frames are not visible in VAA, and hence the prediction error cannot be calculated.
To address this issue, we propose to estimate the prediction error of future frames by forward-backward prediction, and the proposed model is shown in \cref{fig5}.

Moreover, we propose to employ a scene-conditioned auto-encoder to handle the scene-dependent anomalies.
\textcolor{black}{Specifically,} we take the encoding of scene image as the condition of conditional variational auto-encoder (CVAE), train it to generate image features related to the scene, and finally decode the features into the predicted frames.

\subsubsection{Forward-backward Frame Prediction}
As shown in \cref{fig5}, our model includes a forward and a backward frame prediction networks.
The forward network predicts multiple future frames in one shot based on the observed frames, and the backward network reversely predicts an observed frame based on the future frames generated by the forward network and a part of the observed frames.
Our motivation is that, if the future frame is anomalous in forward prediction, the predicted image will be inaccurate.
When we use the inaccurate image as a part of the input for the backward frame prediction model, the output frame will also have a large error with the groundtruth frame, which is available since it has been observed.
Therefore, we can anticipate the future anomalies through the error of forward-backward frame prediction.

At the current time step $t$, the forward network takes the observed frames ${f_{t-n}, \cdots, f_{t-1}}$ as the input, and outputs the predicted frames $\hat{f}_{t}, \cdots, \hat{f}_{t+\alpha}$.
We compute the \textcolor{black}{mean square error (MSE)} loss and L1 loss between every predicted frame $\hat{f}_{t+i}$ ($i \in [0, \alpha]$) and its groundtruth frame \textcolor{black}{to train the forward network}:
\begin{equation}
%	L_f(f_i, \hat{f}_i) = \| f_i - \hat{f}_i \|_2^{2} + \lambda_{L1} |f_i - \hat{f}_i|,
	L_f(f, \hat{f}) = \| f - \hat{f} \|_2^{2} + \lambda_{L1} |f - \hat{f}|,
	\label{eq_Lf}
\end{equation}
where $\lambda_{L1}$ is the weight of L1 loss.

For training the backward network that anticipates the anomaly score of the $i$-th ($i \in [1, \alpha]$) future frame, we feed the predicted future frames $\hat{f}_{t+i}, \cdots,\hat{f}_{t+1}$ \textcolor{black}{and} the real future frames $f_{t+i}, \cdots, f_{t+1}$ respectively along with the observed frames $f_t, \cdots, f_{t+i+1-n}$ into it.
In this way, our backward network can make use of the observed information to make more accurate short-term anomaly anticipation.
The output predicted frames of the two forms of inputs are denoted as $\hat{f}_{t+i-n}^{(1)}$ and $\hat{f}_{t+i-n}^{(2)}$, respectively, which share the same groundtruth frame $f_{t+i-n}$.
We calculate the average MSE and L1 losses between $\hat{f}_{t+i-n}^{(1)}$ and $f_{t+i-n}$, as well as $\hat{f}_{t+i-n}^{(2)}$ and $f_{t+i-n}$ to train the backward network:
\begin{equation}
	 L_b = \frac{1}{2}(L_f(\hat{f}_{t+i-n}^{(1)}, f_{t+i-n}) + L_f(\hat{f}_{t+i-n}^{(2)}, f_{t+i-n})).
\end{equation}

During inference, only the predicted forward future frames $\hat{f}_{t+i}, \cdots,\hat{f}_{t+1}$ and the observed frames $f_t, \cdots, f_{t+i+1-n}$ are required for backward prediction.
For different time steps $t+1, \cdots, t+\alpha$, the backward networks share the same weights.

\subsubsection{Scene-conditioned VAE}
Both the forward and backward networks are three-level U-Nets \cite{UnetConvolutional2015ronneberger} of the same architecture, containing CVAEs that guide the encoding of input frames to be associated with scenes.
The input frames are merged in time and channel dimensions and fed into the encoder of a 2D convolutional network, which outputs three feature maps of different shapes.
The feature maps at $U_2$ and $U_3$ levels are fed into the CVAEs to generate new feature maps conditioned on the scene image.
Then the scene-conditioned feature maps are added to the input of CVAEs with a weight $\gamma \in [0, 1]$.
Finally, the predicted frames are generated through subsequent decoding convolutional layers.

A CAVE takes as input the feature maps of the frames and the encoding of the scene image.
Note that the frames only focus on the local regions \textcolor{black}{of detected objects}, while the objects in the scene image are masked out and only the background is retained.
The scene image is encoded by convolutional layers, concatenated with the frame feature maps and fed into the encoder of CVAE to generate the parameters of a posterior distribution.
We use the reparameterization technique \cite{AutoEncodingVariational2014kingma} to sample \textcolor{black}{latent variables} from the posterior distribution, and feed \textcolor{black}{them} into the CVAE decoder after concatenated with the scene encoding to generate scene-conditioned feature maps.
We assume that the prior distribution is a standard Gaussian distribution and calculate the Kullback-Leibler (KL) divergence between it and the posterior distribution as the loss:
\begin{equation}
	L_{KL}(\mathcal{N}(\hat{\mu}, \hat{\sigma}^2)\|\mathcal{N}(0, 1)) = -\frac{1}{2}(\log \hat{\sigma}^2 - \hat{\mu}^2 - \hat{\sigma}^2 + 1),
\end{equation}
where $\hat{\mu}$ and $\hat{\sigma}^2$ are the mean and variance of the posterior Gaussian distribution.
In testing stage, if the input feature maps do not match the scene, they will be reconstructed by the CVAE with large errors, thereby identifying scene-dependent anomalies.

Finally, the total loss is the sum of the losses of \textcolor{black}{forward prediction, backward prediction}, and KL divergence with the weight of $\lambda_{KL}$.
\textcolor{black}{We minimize the total loss to jointly train the whole model.}

\subsubsection{Anomaly Score}
\label{sec_anomaly_score}
During inference, we calculate the error between the predicted forward frame $\hat{f}_t$ and its groundtruth frame $f_t$ by \cref{eq_Lf} as the anomaly score for VAD:
\begin{equation}
	s(t) = L_f(f_t, \hat{f}_t) .
\end{equation}

For VAA with the anticipation time of $\alpha$, we first estimate the anomaly score of $f_{t+i}$ ($i \in [1, \alpha]$) through forward-backward prediction.
Then, the maximum error in the period of $[t+1, t+\alpha]$ is taken as the anticipation anomaly score:

\begin{equation}
	s(t+1:t+\alpha) = \max(\{L_f(f_{t+i-n}, \hat{f}_{t+i-n})\}_{i=1}^{\alpha}).
	\label{eq_ant_scr}
\end{equation}

Consequently, we can detect and anticipate anomalies simultaneously.

\section{Experiments}

\subsection{Experimental Setup}

\begin{table}[!t]
	\centering
	\caption{Comparison of different methods on the ShanghaiTech, CUHK Avenue, IITB Corridor and NWPU Campus datasets in AUC (\%) metric. \textcolor{black}{The best result on each dataset is shown in bold.}}
	\setlength\tabcolsep{3.9pt}
	\begin{tabular}{@{}llcccc@{}}
		\toprule
		Method      & Year      & ST   & Ave & Cor & Cam \\ \midrule
		FFP\cite{FutureFrame2018liu}         & CVPR 18   & 72.8 & 84.9   & 64.7     & -      \\ \midrule
		MemAE\cite{MemorizingNormality2019gong} 	& ICCV 19   & 71.2 & 83.3   & -        & 61.9   \\
		MPED-RNN\cite{LearningRegularity2019morais}    & CVPR19    & 73.4 & -      & 64.3     & -      \\ \midrule
		MTP\cite{MultitimescaleTrajectory2020rodrigues}         & WACV 20   & 76.0 & 82.9   & 67.1     & -      \\
		VEC-AM\cite{ClozeTest2020yu}      & ACM MM 20 & 74.8 & 89.6   & -        & -      \\
		CDDA\cite{ClusteringDriven2020chang}        & ECCV 20   & 73.3 & 86.0   & -        & -      \\
		BMAN\cite{BMANBidirectional2020lee}        & TIP 20    & 76.2 & 90.0   & -        & -      \\
		Ada-Net\cite{LearningNormal2020song}     & TMM 20    & 70.0 & 89.2   & -        & -      \\
		MNAD\cite{LearningMemoryGuided2020park}        & CVPR 20   & 70.5 & 88.5   & -        & 62.5   \\
		OG-Net\cite{OldGold2020zaheer}      & CVPR 20    & -    & -      & -        & 62.5   \\ \midrule
		CT-D2GAN\cite{ConvolutionalTransformer2021feng}    & ACM MM 21 & 77.7 & 85.9   & -        & -      \\
		ROADMAP\cite{RobustUnsupervised2021wang}     & TNNLS 21  & 76.6 & 88.3   & -        & -      \\
		MESDnet\cite{MultiEncoderEffective2021fang}     & TMM 21    & 73.2 & 86.3   & -        & -      \\
		AMMC-Net\cite{AppearanceMotionMemory2021caia}    & AAAI 21   & 73.7 & 86.6   & -        & 64.5   \\
		MPN\cite{LearningNormal2021lv}         & CVPR 21   & 73.8 & 89.5   & -        & 64.4   \\
		HF\textsuperscript{2}-VAD\cite{HybridVideo2021liu}     & ICCV 21   & 76.2 & \textbf{91.1}   & -        & 63.7   \\ \midrule
		SSAGAN\cite{SelfSupervisedAttentive2022huang}      & TNNLS 22  & 74.3 & 88.8   & -        & -      \\
		DLAN-AC\cite{DynamicLocal2022yang}     & ECCV 22   & 74.7 & 89.9   & -        & -      \\
		LLSH\cite{LearnableLocalitySensitive2022lu}        & TCSVT 22  & 77.6 & 87.4   & 73.5     & 62.2   \\
		VABD\cite{VariationalAbnormal2022li}        & TIP 22    & 78.2 & 86.6   & 72.2     & -      \\ \midrule
		Ours        & -         & \textbf{79.2} & 86.8   & \textbf{73.6}     & \textbf{68.2}   \\ \bottomrule
	\end{tabular}
	\label{exptab1}
\end{table}

\textbf{Datasets.}
We experiment on the ShanghaiTech \cite{RevisitSparse2017luo}, CUHK Avenue \cite{AbnormalEvent2013lu}, IITB Corridor \cite{MultitimescaleTrajectory2020rodrigues} and our proposed NWPU Campus datasets, which are described in \cref{tab1} and the Related Work section.
Our dataset is available at: (it will be released after the double-blind review).
For convenience, we abbreviate the above datasets to "ST", "Ave", "Cor", and "Cam" respectively in the following tables.

\textbf{Evaluation Metric.}
We use the area under the curve (AUC) of receiver operating characteristic (ROC) to evaluate the performance for both VAD and VAA.
Note that we concatenate all the frames in a dataset and then compute the overall frame-level AUC, which is widely adopted.

\textbf{Implementation Details.}
The input frames of our model are the regions of 256$\times$256 pixels centered on objects that detected \textcolor{black}{by} the pre-trained ByteTrack \cite{ByteTrackMultiobject2022zhang} implemented by MMTracking \cite{mmtrack2020}.
For the forward and backward networks, they both take $T_{in}$=8 frames as the input, while they output $T_{out}$=7 and $T_{out}$=1 frames, respectively.
The 1st frame output by the forward network is used for anomaly detection, and the 2nd to 7th frames are fed into the backward network for anomaly anticipations of different anticipation times.
%We design the U-Net encoder based on ResNet [], and the dimensions ($C$, $H$$\times$$W$) of the feature maps at $U_1$, $U_2$ and $U_3$ levels are [64, 128\textsuperscript{2}], [64, 64\textsuperscript{2}], [128, 32\textsuperscript{2}] respectively.
We design the encoder of U-Net based on ResNet \cite{DeepResidual2016he} and the decoder are multiple convolutional layers.
The network for scene encoding is a classification model to classify scenes, which is firstly trained with known scene information, and then frozen during training the entire model.
%The dimension of the latent variable $\mathbf{z}$ in each CVAE is 2.
The weights $\gamma$, $\lambda_{L1}$ and $\lambda_{KL}$ are 1, 1 and 0.1 by default.
We adopt the maximum local error \cite{ContextRecovery2022caoa} to focus on the errors in local regions.
\textcolor{black}{Please refer to the supplementary material for a detailed description of our model.}

\subsection{VAD Performance Benchmarking}

The comparison between our method and other existing methods on the ShanghaiTech, CUHK Avenue, IITB Corridor and NWPU Campus datasets is shown in \cref{exptab1}.
We reproduce a total of 7 recent \textcolor{black}{reconstruction-based \cite{OldGold2020zaheer}, distance-based \cite{LearnableLocalitySensitive2022lu} and prediction-based \cite{MemorizingNormality2019gong, LearningMemoryGuided2020park, AppearanceMotionMemory2021caia, LearningNormal2021lv, HybridVideo2021liu}} methods on our NWPU Campus dataset \textcolor{black}{using their official codes}.
For a fair comparison, the self-supervised learning based methods \cite{ClusterAttention2020wang, AnomalyDetection2021georgescu, VideoAnomaly2022wang} are excluded, and we use the same detected objects as the inputs for the reproduced methods.
The $\gamma$ in our model is set to 0 for those datasets without scene-dependent anomalies.
As can be seen in \cref{exptab1}, our method outperforms the others on the NWPU Campus, IITB Corridor and ShanghaiTech datasets, all of which contain over 10 classes of abnormal events.
The superior performance demonstrates the advantage of our method for complex and large-scale VAD.
We find that the relatively low performance on the CUHK Avenue is mainly due to the inaccurate object tracking of the tracking algorithm, which is caused by the low resolution of this dataset.
The performance for VAD on the NWPU Campus is lower than that on other datasets because our dataset contains various types of anomalies, and each anomaly has multiple manifestations, making it much more challenging than other datasets.

\subsection{Study on Scene-dependent Anomalies}

\begin{table}[!t]
	\centering
	\caption{AUCs (\%) of different methods on scene-dependent anomalous datasets. The ShanghaiTech-sd dataset used in this table is reorganized by us. \textcolor{black}{The best results are shown in bold.}}
	\begin{tabular}{@{}lcc@{}}
		\toprule
		Method           & Cam  & \begin{tabular}[c]{@{}c@{}}ST-sd\\(reorganized)\end{tabular} 	   \\ \midrule
		MemAE \cite{MemorizingNormality2019gong}		 & 61.9 & 67.4                                                                \\
		MNAD \cite{LearningMemoryGuided2020park}            & 62.5 & 68.2                                                             \\
		OG-Net \cite{OldGold2020zaheer}          & 62.5 & 69.6                                                             \\
		AMMC-Net \cite{AppearanceMotionMemory2021caia}        & 64.5 & 64.9                                                             \\
		MPN \cite{LearningNormal2021lv}             & 64.4 & 76.9                                                             \\
		HF\textsuperscript{2}-VAD \cite{HybridVideo2021liu}         & 63.7 & 70.8                                                             \\ \midrule
		Ours ($\gamma$=0)  & 65.8 & 70.4                                                             \\
		Ours ($\gamma$=1)   & \textbf{68.2} & \textbf{82.7}                                                             \\ \bottomrule
	\end{tabular}
	\label{exptab2}
\end{table}

\begin{figure}[!t]
	\centering
	\includegraphics[]{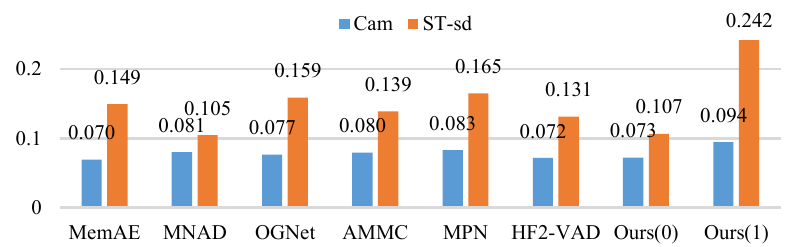}
	\caption{\textcolor{black}{Score gaps of different methods. "Ours (0)" and "Ours (1)" denote our methods with $\gamma=0$ and $\gamma=1$, respectively. A higher value means better.}}
	\label{score_gap_sd}
\end{figure}

%To study scene-dependent anomaly detection, we not only experiment on our NWPU Campus dataset, but also reorganize a new dataset named ShanghaiTech-sd using a part of the videos from the ShanghaiTech dataset.
In addition to our NWPU Campus dataset, we also reorganize a new dataset named ShanghaiTech-sd using a part of the videos from the ShanghaiTech dataset to specifically study scene-dependent anomaly detection.
ShanghaiTech-sd contains 4 scenes where "cycling" is set as a scene-dependent anomaly.
%ShanghaiTech-sd contains 4 scenes, \ie, "01", "06", "10" and "12", in which the "01" scene contains training videos of "cycling" from the original testing set.
%In other scenes, there are only walking in the training videos.
%Therefore, cycling is normal in the "01" scene but abnormal in other scenes, which makes it a kind of scene-dependent anomalous event.
%The performances of \textcolor{black}{different} methods for scene-dependent anomaly detection are shown in \cref{exptab2}.
The performances of \textcolor{black}{different methods are} shown in \cref{exptab2}.
It can be seen that the proposed scene-conditioned VAE (\ie $\gamma$=1) makes a significant improvement, with increases of 2.4\% and 12.3\% on the NWPU Campus and ShanghaiTech-sd, respectively, surpassing other methods by a margin.
\textcolor{black}{We analyze the score gaps between normal and abnormal scores of those methods, as can be seen in \cref{score_gap_sd}.
In particular, the score gap of our method with $\gamma$=1 is obviously higher than that with $\gamma$=0 and other methods, suggesting that the proposed scene-conditioned VAE can distinguish scene-dependent anomalies.}
We provide the \textcolor{black}{details} of the ShanghaiTech-sd dataset and more analysis in the supplementary material.

\subsection{Video Anomaly Anticipation}

\begin{table}[!t]
	\centering
	\caption{AUCs (\%) for video anomaly anticipation with different anticipation times (\ie $\alpha_{t}$ seconds) on the NWPU Campus dataset. "f" and "b" denote forward and backward predictions.}
	\begin{tabular}{@{}lcccccc@{}}
		\toprule
		~~$\alpha_{t}$ & 0.5s & 1.0s & 1.5s & 2.0s & 2.5s & 3.0s \\ \midrule
		Chance            & 50.0 & 50.0 & 50.0 & 50.0 & 50.0 & 50.0 \\
%		Human             & \multicolumn{6}{c}{90.4}                \\ \cmidrule(l){2-7} 
		Human             & - & - & - & - & - & 90.4  \\
		Ours (f-only)       & 65.2 & 64.6 & 64.2 & 63.6 & 63.1 & 62.5 \\
		Ours (f+b)             & 65.8 & 65.3 & 64.9 & 64.6 & 64.2 & 64.0 \\ \bottomrule
	\end{tabular}
	\label{exptab3}
\end{table}

We conduct experiments on the NWPU Campus dataset for VAA with different anticipation times, as shown in \cref{exptab3}.
We report the results of stochastic anticipations ("Chance") and human beings ("Human").
Four volunteers not involved in the construction of the dataset participate in the evaluation of anomaly anticipation.
Since humans cannot perceive time precisely, the volunteers only anticipate whether an anomalous event will occur in 3 seconds or not.
The result of "Human" is the average performance of all the volunteers.
%For the forward-only model (\ie, f-only), we take the error between the predicted future frame after $\alpha_{t}$ seconds and the last observed frame as the anticipated anomaly score.
For the forward-only model (\ie, f-only), we calculate the maximum error between the predicted future frames in $\alpha_{t}$ seconds and the current frame, which is then taken as the anticipated anomaly score.
The forward-backward model (\ie, f+b) computes anomaly scores as mentioned in \cref{sec_anomaly_score}.
It can be seen that our forward-backward prediction method is more effective than the forward-only method.
However, there is still much room for improvement compared with the performance of humans, which demonstrates that the proposed dataset and VAA task are extremely challenging for algorithms.

\section{Conclusion}
In this work, we propose a new comprehensive dataset NWPU Campus, which is the largest one in semi-supervised VAD, the only one considering scene-dependent anomalies, and the first one proposed for video anomaly anticipation (VAA).
We define VAA to anticipate whether an anomaly will occur in a future period of time, which is of great significance for early warning of anomalous events.
Moreover, we propose a forward-backward scene-conditioned model for VAD and VAA as well as handling scene-dependent anomalies.
In the future, our research will focus not only on the short-term VAA, but also on long-term anticipation.

\section*{Acknowledgments}
This work is supported by the National Natural Science Foundation of China (Project No. U19B2037, 61906155, 62206221), the Key R\&D Project in Shaanxi Province (Project No. 2023-YBGY-240), the Young Talent Fund of Association for Science and Technology in Shaanxi, China (Project No. 20220117), and the National Key R\&D Program of China (No. 2020AAA0106900).

%%%%%%%%% REFERENCES
{\small
\bibliographystyle{unsrt}
\bibliography{mybib}
}

\end{document}